\documentclass{article}

\usepackage{microtype}
\usepackage{graphicx}
\usepackage{booktabs} 
\usepackage{subcaption}

\usepackage{hyperref}

\usepackage[accepted]{icml2023}

\usepackage{amsmath}
\usepackage{amssymb}
\usepackage{mathtools}
\usepackage{amsthm}

\usepackage[capitalize,noabbrev]{cleveref}

\theoremstyle{plain}

\theoremstyle{definition}

\theoremstyle{remark}

\newcommand{\Comments}{1}
\newcommand{\mynote}[2]{\ifnum\Comments=1\textcolor{#1}{#2}\fi}
\newcommand{\mytodo}[2]{\ifnum\Comments=1%
	\todo[linecolor=#1!80!black,backgroundcolor=#1,bordercolor=#1!80!black]{#2}\fi}

\usepackage[textsize=tiny]{todonotes}

\usepackage{multirow}

\icmltitlerunning{Equivariance Is Not All You Need}

\begin{document}

\twocolumn[
\icmltitle{Equivariance Is Not All You Need: Characterizing the Utility of Equivariant Graph Neural Networks for Particle Physics Tasks}



\icmlsetsymbol{equal}{*}

\begin{icmlauthorlist}
\icmlauthor{Savannah Thais}{cu}
\icmlauthor{Daniel Murnane}{lbl}
\end{icmlauthorlist}

\icmlaffiliation{cu}{Columbia University, Data Science Institute, New York, NY USA}
\icmlaffiliation{lbl}{Lawrence Berkeley National Laboratory, Computing Sciences Research, Berkeley, CA USA}

\icmlcorrespondingauthor{Savannah Thais}{st2565@columbia.edu}

\icmlkeywords{Machine Learning, ICML}

\vskip 0.3in
]



\printAffiliationsAndNotice{\icmlEqualContribution} 

\begin{abstract}
Incorporating inductive biases into ML models is an active area of ML research, especially when ML models are applied to data about the physical world. Equivariant Graph Neural Networks (GNNs) have recently become a popular method for learning from physics data because they directly incorporate the symmetries of the underlying physical system. Drawing from the relevant literature around group equivariant networks, this paper presents a comprehensive evaluation of the proposed benefits of equivariant GNNs by using real-world particle physics reconstruction tasks as an evaluation test-bed. We demonstrate that many of the theoretical benefits generally associated with equivariant networks may not hold for realistic systems and introduce compelling directions for future research that will benefit both the scientific theory of ML and physics applications. 
\end{abstract}

\section{Introduction and Background}
\label{sec:intro}
Over the past several years, Machine Learning (ML) has been established as a core component of many types of physics research \cite{mlphysics1,mlphysics2,mlphysics3}. Because physics is governed by many mathematical symmetries and analytical laws, physics-related applications of ML lend themselves particularly well to the inclusion of inductive biases. 

In particular, equivariant graph neural networks (GNNs) have emerged as a popular approach in many areas of physics research, including particle physics \cite{symmetry_physics_review}, fluid dynamics \cite{fluiddynamics}, molecular physics \cite{nequip}, and materials science \cite{materialsscience}. Equivariant GNNs combine several different types of inductive biases. As explained below, GNNs are permutation equivariant by construction and the graph itself (a combination of nodes and connective edges) incorporates an explicit relational or structural inductive bias into the data representation. Equivariant GNNs add an additional symmetry-based inductive bias by requiring that the function learned by the GNN is equivariant under transformations of some specified symmetry group.

While there are many types of GNNs, we will briefly describe message passing GNNs specifically \cite{messagepassing_gnn}, as they are the kind used in the example experiments discussed later in this paper. Basic message passing GNNs update the representations of graph nodes by exchanging information between neighboring nodes. In each message passing iteration, nodes aggregate information from their neighbors by applying a learnable function to the features $h_j$ of neighboring nodes $x_j$ (possibly as well as the central node $x_i$ and any features of the connecting edges $e_{i,j}$); this transformed neighborhood information is aggregated by a permutation equivariant function to form the `message', which is then combined with the central node's current features to produce an updated representation. This process is described mathematically as \begin{equation} h_i^{l+1}=\psi(h_i^l,\Box_{j\in N(i)}m_{ij}) \label{eq:node_update}\end{equation} where $m_{ij}=\phi(h^l_i,h^l_j,e_{ij})$ is the message function, $\Box$ is the aggregation function, and $\psi$ is the node update function.

An ML model is equivariant if for some symmetry group $G$ with transformations $T_g:X\rightarrow X$ operating on the input space $X$ the model is composed of functions $\phi : X \rightarrow Y$ where there is some transformation $S_g$ on the output space Y such that $\phi(T_g(x))=S_g(\phi(x))$ for all $g\in G$. There are many ways to design a model that is group equivariant by construction. For example, with group convolutions \cite{groupconv} and using higher order group representations to parameterize learnable filters \cite{representation_gnns}. A popular approach to building equivariant GNNs specifically is to build the messages only from group equivariant vectors and invariant scalars \cite{egnn, villar2021scalars}. In this approach, the position features of the nodes, $x_i$ (where $x_i^0$ is the initial spatial position of the node), are updated separately from the other node features. Thus, in addition to Equation \ref{eq:node_update}, we have  \begin{equation} x_i^{l+1}=x_i^l+C\Box_{j\in N(i)}\rho(x_i,x_j,m_{ij}) \label{eq:spatial_update}\end{equation} where $\rho$ is equivariant under the symmetry group transformation and $C$ is a scalar weight.  

As discussed in Section \ref{sec:equivariance_benefits}, there are many proposed benefits to using equivariant networks (of which equivariant GNNs, the focus of this paper, are a subclass) to model symmetry-obeying systems. In this paper, using particle physics tasks as a test-bed, we seek to characterize whether these benefits actually manifest when equivariant models are applied to real-world physical systems. A better understanding of what types of inductive biases yield accurate and efficient models is critical to enabling a wide spectrum of physics research and experiments. \footnote{Note that the scrutiny in this work is directed towards equivariance in particle physics, and the conclusions may not apply directly to systems with discrete symmetries or significantly different metrics, as in chemistry and biology.}

\section{Benefits of Equivariant Models}
\label{sec:equivariance_benefits}
Group equivariant networks are a popular and active area of ML research. This work is often motivated both by the success of models with certain equivariant properties, such as translation symmetry in Convolutional Neural Networks (CNNs) on the domain of natural images with its approximately translation symmetry, and by the abundance of symmetries inherent to the physical world (and consequently, a plethora of datasets drawn from astronomy, physics, and chemistry). Researchers have suggested and, in some cases, demonstrated a variety of potential benefits, including:

\paragraph{Accuracy}
The equivariant models introduced in Section \ref{sec:intro} and many other proposed equivariant architectures achieve state of the art (SotA) accuracy on the selected benchmark data sets they are evaluated on for publication. This makes sense as SotA performance is typically an unstated criteria for successful publication in ML venues. 

Indeed, improvements in model accuracy are often cited as the main motivation for studying equivariant architectures. The performance successes of CNNs and GNNs broadly are often attributed to their respective inherent translation and permutation equivariance. This narrative is repeated for group equivariant networks to foster an expectation of improved model performance; see, for example, the introduction sections of \cite{egnn} and \cite{villar2021scalars}.

\paragraph{Generalizability}
In theory, a group equivariant network should be able to learn a complete orbit (i.e. the manifold formed by the action of the symmetry group) from a single sample. The network would then be able to effectively model any other input from the same orbit. This phenomenon is called generalization and we refer to a model's capacity to generalize as generalizability. Presumably, by enforcing equivariance, the network is able to learn precise, reduced representations of the data domain, which aids in generalization.

Some studies of equivariant networks, though not all, include tests that demonstrate the model's ability to generalize under group transformations (e.g. Figure 3 in \cite{lorentz_net}). Furthermore, some research takes up the task of generalization specifically. For example, the authors of \cite{generalization} demonstrated that their E(2)-equivariant CNN with additional modifications for robustness to scale changes generalizes better under translations, rotations, and linear scaling than a non-equivariant model (although it still does not generalize across the entire test data domain).

\paragraph{Model Efficiency}
Previous work has suggested that with symmetries of the data built into a model, equivariant networks have an `easier' time learning the optimal function to minimize the training loss. Concretely, it has been proposed that for a given number of learnable parameters, an equivariant model may be able to use those parameters more efficiently than a non-equivariant model. In practice, this model efficiency was demonstrated by the authors of \cite{cohen2017steerable}, who showed that a `steerable CNN' that creates representations through a composition of elementary features, each associated with a particular kind of symmetry, was able to utilize model parameters more efficiently to yield a smaller SotA image classifier for the CIFAR benchmark. 

Model efficiency is highly desirable, particularly in resource constrained computing environments. Consequently, some authors, for example \cite{lgn}, cite the reduced size of their equivariant models compared to similarly performing non-equivariant models as the main benefit, even when they may not achieve full SotA accuracy. 

\paragraph{Data Efficiency}
The `traditional' method of training an unconstrained model to respect a symmetry is through data augmentation, which often substantially increases the size of the required training data set, and therefore the computing resources needed to train the model. It is expected that because an equivariant network is constrained to only learn symmetry-obeying functions, that they would need fewer training examples to achieve similar performance; put another way, they would demonstrate improved data efficiency. 

The authors of \cite{egnn} demonstrate the data efficiency of their E(3)-equivariant GNN by comparing model performance as a function of number of samples in the training data set with a non-equivariant model. The equivariant GNN performs notably better than the unconstrained model in the small data regime; interestingly, it also performs better better than another equivariant model in the large data regime. Additionally, the authors of \cite{sample_complexity} provide theoretical evidence that equivariant models demonstrate better data efficiency than unconstrained models, at least in the case of enforcing equivariance through the use of invariant kernels.

We also note that model interpretability or explainability is often suggested as a benefit to building equivariant networks \cite{symmetry_physics_review}. However, since there is no clear metric for evaluating the interpretability or explainability capacity of a model, particularly for the physics data sets studied in this paper, we do not evaluate that claim in this paper.

\section{Symmetries and Machine Learning for Physics}
\label{sec:for_physics}
Although the benefits of equivariance described in Section \ref{sec:equivariance_benefits} have been demonstrated for certain combinations of model type, symmetry, and task, none of them are supported by strong general theoretical guarantees or by all of the experimental evidence. In order to better characterize the possible benefits of enforcing equivariance in ML models, we investigate the performance of equivariant GNNs on two important particle physics data reconstruction tasks.

\subsection{Top Jet Tagging}
\begin{figure}[h]
\includegraphics[width=8.5cm]{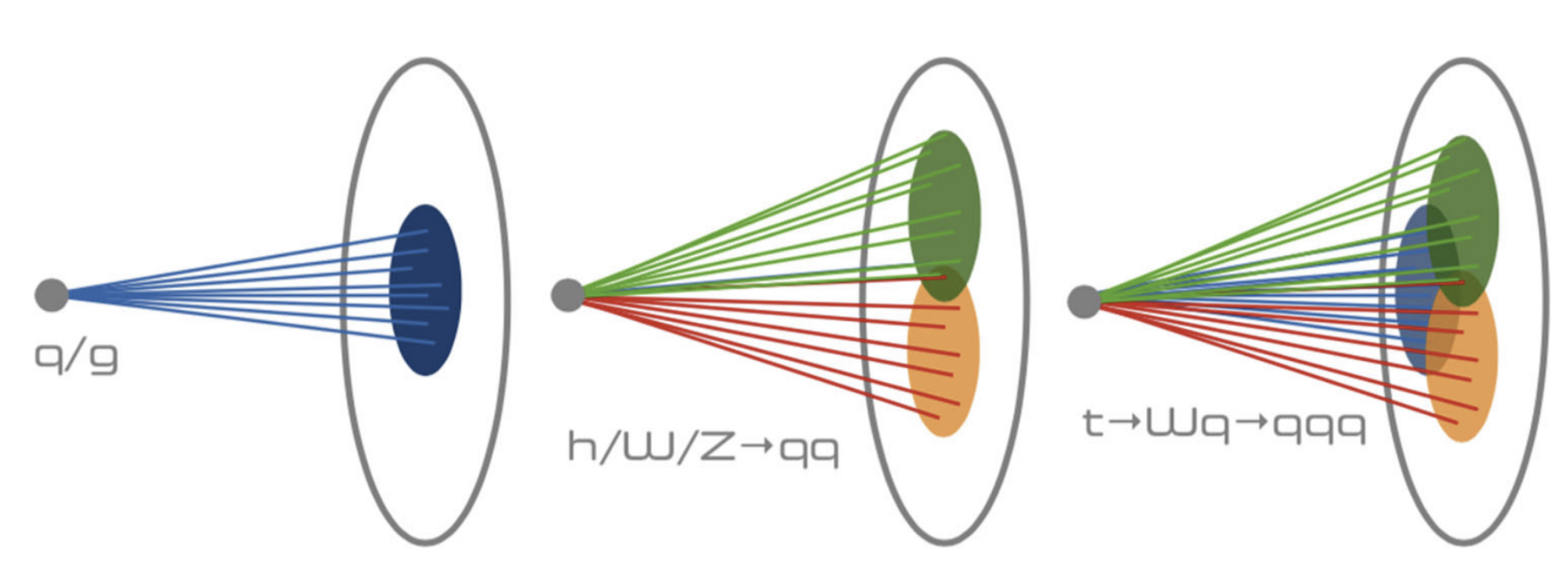}
\caption{Pictorial representation of three common types of jets: a single quark/gluon jet (left),a heavy jet decaying to two quarks (center), and a high momentum top quark to heavy boson to quarks decay chain.  Figure reproduced from \cite{jedinet}.\label{fig:jets}}
\end{figure}

In high-energy particle physics experiments, like the ATLAS and CMS detectors at the Large Hadron Collider (LHC), jets play a pivotal role in the study of particle interactions. Jets are collimated sprays of particles formed when energetic quarks or gluons are produced during proton-proton collisions. Due to the nature of the strong force and the confinement of quarks and gluons, these particles rapidly fragment and hadronize, resulting in a collection of detectable particles clustered together within a narrow cone (Figure \ref{fig:jets}). Identifying the type of originating quark or gluon (`jet tagging') is essential to enable precision measurements and searches for new physics processes. 

Lorentz symmetry is a fundamental principle in physics that arises from the theory of special relativity; it describes the invariance of physical laws under rotations and boosts in spacetime. The Lorentz group is the set of all matrices that preserve the bilinear form $\eta(u,v)=u^TJv$ for spacetime vectors $u,v\in\mathbb{R}^4$ and $J=diag(1,-1,-1,1)$, the Minkowski metric. Lorentz symmetry, with inertial frames restricted to be positively oriented and positively time-oriented, is described by the proper orthochronous Lorentz group, $\mathrm{SO(1,3)}^+$. The task of top jet tagging is invariant under Lorentz transformations as the originating quark or gluon type does not depend on the rotation or boost of the jet.

Top jet tagging is a popular benchmarking task for studying ML models in particle physics. In particular, several Lorentz equivariant top jet taggers have recently been proposed. The Lorentz Group Equivariant Neural Network (LGN) leverages higher order tensor products of the Lorentz group to construct equivariant learnable linear operations \cite{lgn}. LorentzNet modifies the standard GNN message construction equation to contain only equivariant information by leveraging the Minkowski dot product \cite{lorentz_net}. Concretely, Equation \ref{eq:node_update} becomes $x_i'=x_i+c\Box_{j\in N}\phi_x(m_{i,j}\cdot x_j)$ and the message becomes \begin{equation}
m_{i,j}=\phi(h_i,h_j, ||x_i-x_j||^2,\langle x_i,x_j\rangle) 
\end{equation}
with $\langle x_i,x_j\rangle$, the Minkowski dot product. VecNet similarly enforces Lorentz equivariance by constraining the GNN message, however it also includes a separate unconstrained `standard' GNN message construction that is merged with the equivariant message before the final fully connected layers of the GNN; the size of this each message passing network can be adjusted, allowing interpolation between a fully equivariant and fully unconstrained GNN \cite{vecnet}. PELICAN uses the complete set of Lorentz invariants as inputs to the GNN and uses all 15 of the permutation invariant transformations from rank 2 tensors to rank 2 tensors as aggregators \cite{pelican}.

In order to understand the impact of enforcing Lorentz equivariance in top jet tagging ML models, we compare the performance of these equivariant models to several other popular taggers that achieve SotA or near SotA performance. EFP is a binary linear classifier that uses Energy Flow Polynomials, a complete set of jet substructure observables which form a discrete linear basis for all infrared- and collinear-safe observables, as input features \cite{efp}. ResNeXt \cite{resnext} is a very deep two-dimensional CNN that uses jet images \cite{jet_images} as inputs rather than graph-based jet representations (the specific implementation of ResNeXt for top jet tagging is described in \cite{particlenet}). ParticleNet and Particle Flow Network (PFN) both treat the jet as an unordered set of particles; ParticleNet is a graph convolution GNN architecture with dynamic graph updates \cite{particlenet} while PFN is based on the DeepSets framework \cite{pfn}. 

The performance of these models are summarized in Table \ref{tab:topjet_results} (note we include performance metrics for two different configurations of VecNet, one with higher accuracy and one with higher model efficiency, as reported in the original paper). All models are evaluated on the public Top Quark Tagging Reference Dataset \cite{top_dataset} which consists of 2M proton-proton collision events containing either a hadronically decaying top quark (signal) or QCD dijets (background).

\begin{table*}[!ht]
\centering
\begin{tabular}{@{}lccccc@{}}
\toprule
Model       & Accuracy                         & AUC                              & $\epsilon_B^{-1}$ & $\mathrm{N}_{\mathrm{parameters}}$  & Ant Factor v2 ($\times 10^5$)    \\ \midrule
ResNeXt     & 0.936                            & 0.984                            & $1122 \pm 47$     & 1.46 M                             & 4.28  \\
ParticleNet & 0.938                            & 0.985                            & $1298 \pm 46$     & 498k                                     & 13.4  \\
PFN         & 0.932                            & 0.982                            & $891 \pm 18$      & 82k                                 & 67.8  \\
EFP         & 0.932                            & 0.980                            & 384               & 1k                        & \textbf{5000}  \\
LGN         & 0.929                            & 0.964                            & $435  \pm 95$     & 4.5k                                & 617   \\
VecNet.1    & 0.935                            & 0.984                            & - -               & 633k                               & 9.87  \\
VecNet.2    & 0.931                            & 0.981   & - -                   & 15k                                           & 350   \\
PELICAN     & \textbf{0.943}                  & \textbf{0.987}                  & \textbf{2289} $\pm$ \textbf{204}    & 45k                                 & 171   \\
LorentzNet  & 0.942                            & 0.9868                           & $2195 \pm 173$    & 220k                                       & 35.0  \\ \bottomrule
\end{tabular}
\caption{Performance comparison of SotA and near-SotA top tagging models\footnotemark. VecNet performance metrics are reproduced from \cite{vecnet} while all other values are reproduced from \cite{pelican}. $\epsilon_B^{-1}$ is the background rejection rate (the inverse of the false postive rate) for 30\% signal efficiency. Ant Factor v2 is $[(1-AUC)\cdot\mathrm{N}_{\mathrm{parameters}}]^{-1}.$ \label{tab:topjet_results}}
\end{table*}

\subsection{Charged Particle Tracking}
\begin{figure}[h]
\centering
\includegraphics[width=7.5cm]{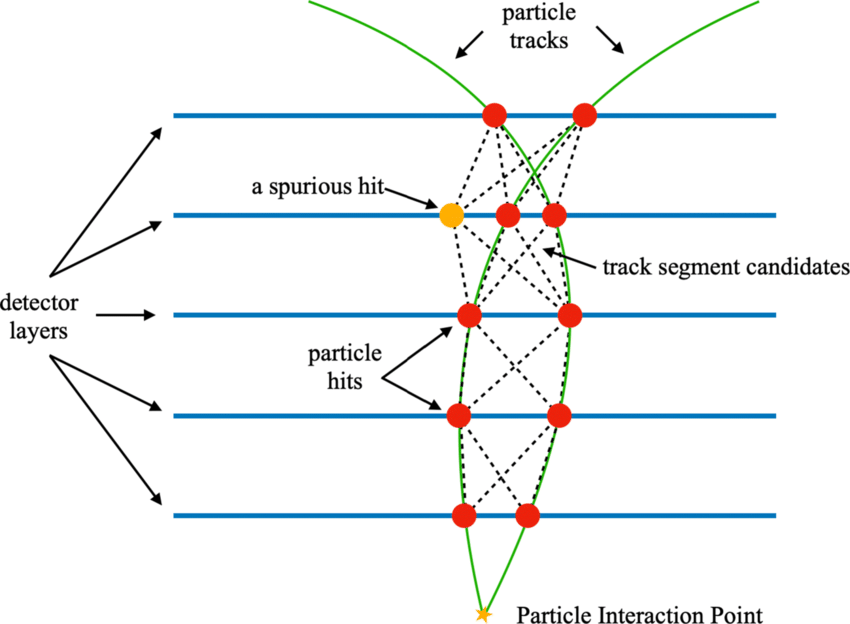}
\caption{Drawing of particle track reconstruction problem. Particles interact near the origin of the coordinate system and travel outward, bending in a direction depending on their magnetic charge. Figure reproduced from \cite{track_diagram_source}.\label{fig:tracks}}
\end{figure}

When charged particles produced in proton-proton collisions at the LHC traverse the experiments they leave energy deposits (`hits') in different layers of the detectors (Figure \ref{fig:tracks}). Identifying which hits belong to the trajectory (`track') of an individual particle (`tracking') and extracting the kinematic properties of each trajectory is crucial for understanding particle interactions, identifying new particles, and studying their properties. Tracking is a challenging task due to high particle mutliplicities resulting in many overlapping tracks; it is also the most computationally intensive reconstruction task at the ATLAS and CMS experiments. Recently, edge-classifying GNNs have emerged as a popular approach for reducing the computational burden of tracking, particularly for planned upgrades of the LHC \cite{gnn_tracking_overview}. Tracking GNNs typically operate on a graph with track hits as nodes and possible track segments (usually constructed through a physics-motivated spatial proximity metric) as edges; the GNN then classifies track segments as true (part of a real particle trajectory) or false (spuriously created during the graph construction) and a final clustering or walk-through algorithm links the positively identified track segments into full tracks.

Equivariant ML-based tracking models have not been well-studied, in part because although the task of particle tracking is Lorentz equivariant, the popular benchmark data set, TrackML, does not contain the timing information necessary to train Lorentz equivariant models. However, a recent paper presented EuclidNet, an SO(2) equivariant GNN developed for charged particle tracking \cite{euclidnet}. Although physics is symmetric under 3-dimensional rotations, the authors focus on SO(2) symmetry in the plane perpendicular to the beamline of a particle accelerator, rather than full SO(3) symmetry, due to the cylindrical construction of detector experiments at the LHC. EuclidNet uses message construction 
\begin{equation}
m_{i,j}=\phi(h_i,h_j, ||x_i-x_j||^2)     
\end{equation} as input to Equations \ref{eq:node_update} and \ref{eq:spatial_update}.

The performance of EuclidNet and a SotA non-equivariant tracking GNN (InteractionNet \cite{interactionnet} at various sizes are presented in Table \ref{tab:tracking_results}. All models are evaluated on the TrackML data set \cite{trackml}, a public data set consisting of a set of simulated proton-proton  collision  events, each containing three-dimensional  hit positions  and  truth  information  about  the  particles  that  generated  them. The events are simulated with 200 pileup, simulating the high particle multiplicity expected in future upgrades of the LHC. 

\begin{table*}[!ht]
\centering
\begin{tabular}{lcccc}
\hline
Model          & $\mathrm{N}_{\mathrm{hidden}}$ & AUC                           & $\mathrm{N}_{\mathrm{parameters}}$ & Ant Factor v2 ($x10^5$) \\ \hline
EuclidNet      & \multirow{2}{*}{8}             & \boldmath{$0.9913 \pm 0.004$} & 967                                & \textbf{11887}          \\
InteractionNet &                                & $0.9849 \pm 0.006$            & 1432                               & 4625                    \\ \hline
EuclidNet      & \multirow{2}{*}{16}            & $0.9932 \pm 0.003$            & 2580                               & \textbf{5700}           \\
InteractionNet &                                & $0.9932 \pm 0.004$            & 4392                               & 3348                    \\ \hline
EuclidNet      & \multirow{2}{*}{32}            & $0.9941 \pm 0.003$            & 4448                               & 3810                    \\
InteractionNet &                                & \boldmath$0.9978 \pm 0.003$   & 6448                               & \textbf{7049}           \\ \hline
\end{tabular}
\caption{Performance comparison between EuclidNet and InteractionNet on the TrackML dataset. The results for both models are averaged over 5 runs. Values are reproduced from \cite{euclidnet}}\label{tab:tracking_results}
\end{table*}

\section{Evaluating Equivariance}
\label{sec:evaluating}
Using the performance of these top jet tagging and tracking models, we now evaluate whether the proposed benefits of equivariance described in Section \ref{sec:equivariance_benefits} are realized. 

\subsection{Accuracy}
At first glance, it does seem that enforcing equivariance improves model accuracy. Amongst the top jet taggers, the two models with the highest accuracy and AUC, PELICAN and LorentzNet, are both Lorentz invariant. However, another Lorentz equivariant model, LGN, has the lowest accuracy and AUC amongst all models considered, indicating that the accuracy improvements may be due, at least in part, to other architecture choices. In particular, PELICAN takes all possible Lorentz invariants as input features and innovatively utilizes 15 different aggregators, creating a richer data representation within the model. Furthermore, VecNet.1 outperforms both LGN and several of the non-equivariant models, despite only being partially equivariant. 

For tracking models, the picture is similarly unclear. For very small model sizes, the SO(2) equivariant EuclidNet has a higher AUC. However, as the size of the model latent space increases, the EuclidNet performance begins to plateau and the unconstrained InteractionNet becomes more performant. We explore the relationship between model size and accuracy further in Section \ref{sec:model_eff}.

\footnotetext{We do not include background rejection values for VecNet.1 and VecNet.2 because the authors have noted an error in the original calculation but have not yet published corrected numbers.}

\subsection{Generalizability}
\begin{figure}[htb!]
\centering
\begin{subfigure}[b]{0.55\textwidth}
   \includegraphics[width=8cm]{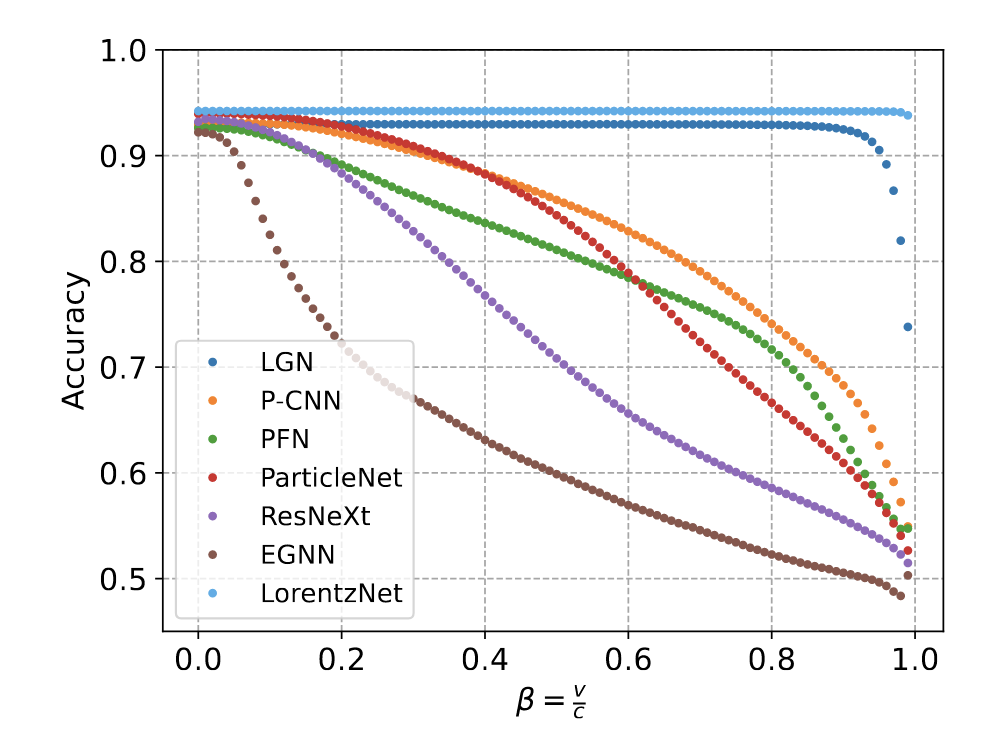}
   \caption{}
   \label{fig:lorentznet_test} 
\end{subfigure}

\begin{subfigure}[b]{0.55\textwidth}
   \includegraphics[width=8cm]{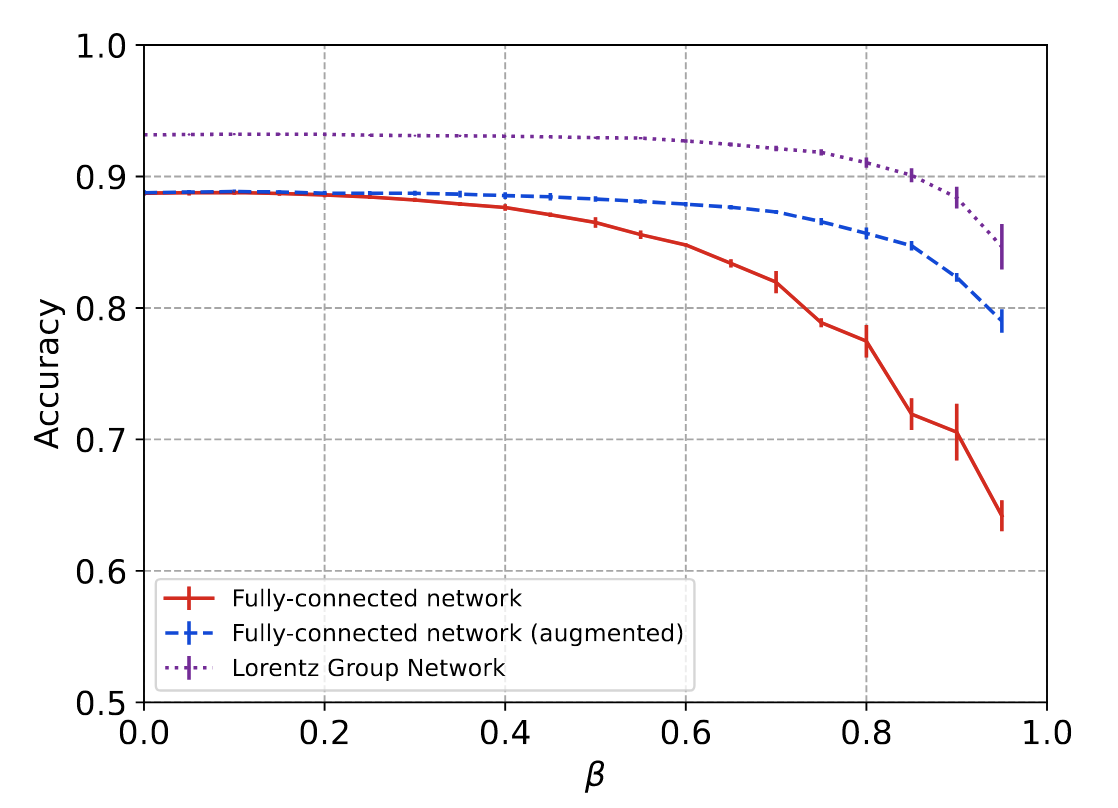}
   \caption{}
   \label{fig:equipaper_test}
\end{subfigure}

\caption{(a) Accuracy of top jet tagging models under a range of Lorentz boosts (parameterized by $\beta$). Reproduced from \cite{lorentz_net}. (b) The accuracy of various neural networks for top jet tagging under a range of Lorentz boosts (parameterized by $\beta$). Reproduced from \cite{symmetry_physics_review}}
\end{figure}

In order to describe the generalizability of a model to transformations of the input data, one can test how the model accuracy changes as continuous transformations are applied. Figure \ref{fig:lorentznet_test} shows the accuracy of top jet taggers as a function of the Lorentz boost applied to the data reference frame, parameterized by $\beta=v/c$ for speed of light $c$. The accuracies of the non-Lorentz-equivariant models decline quickly as boosts are applied, while the accuracy of LorentzNet remains uniform for almost the entire range of boosts, only declining slightly when the largest boost is applied. This indicates that, as expected, LorentzNet is able to generalize well. 

Interestingly, however, another Lorentz Equivariant model, LGN, does not generalize quite as well, as the accuracy begins to decline steeply for large Lorentz boosts, ending at only ~70\% of its original accuracy when the largest boost is applied. In the original paper, the authors attribute this decrease in performance to instabilities arising from the bit precision used to store the model weights \cite{lgn}, indicating that the particular approach to enforcing equivariance in an ML architecture has an impact on the model's ability to fully generalize.

\begin{figure}[htb!]
    \centering
    \includegraphics[width=8cm]{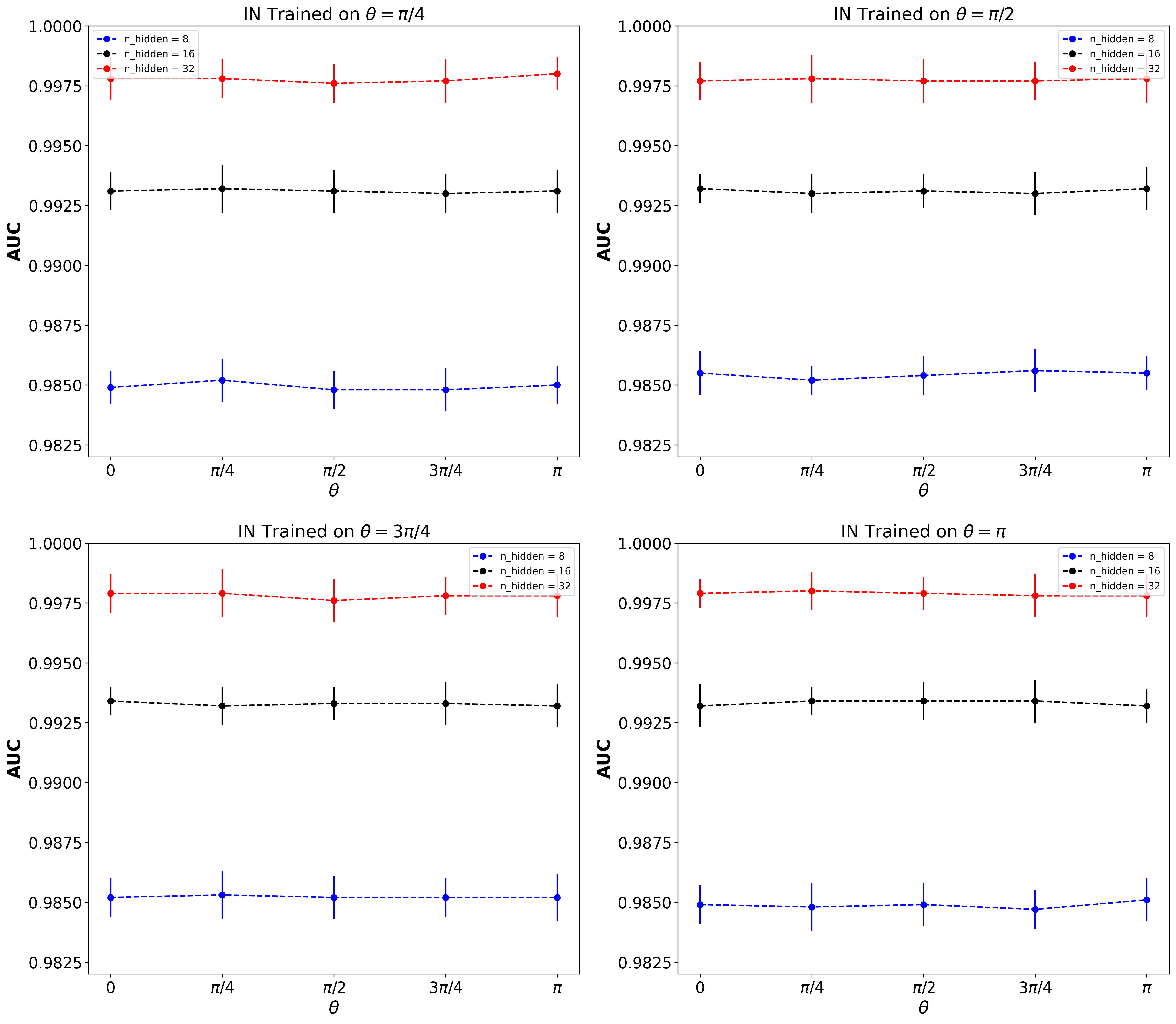}
  \caption{The AUC at inference time plotted as a function of rotations in the input space by an angle $\theta$ for models trained on a set of rotations of the dataset by $\theta \in [0, \pi/4, \pi/2, 3\pi/4, \pi]$ for InteractionNet, calculated over 5 independent inference runs. Reproduced from \cite{euclidnet}.}
  \label{fig:rotation_tests}
\end{figure}

Of course, equivariance is not the only way for a model to learn to generalize under a symmetry transformation. As shown in Figure \ref{fig:equipaper_test}, a simple 4-layer fully-connected neural network trained on an augmented version of the Top Jet Tagging Dataset with random Lorentz boosts applied to training events is able to generalize equally as well as LGN (once the difference in accuracy in non-boosted data is accounted for). Additionally, in some cases a model is able to learn robustness to transformations even without training data augmentation. As shown in Figure \ref{fig:rotation_tests}, the non-equivariant InteractionNet maintains stable performance across a range of 2-dimensional rotations applied to the testing data set. 

\subsection{Model Efficiency}
\label{sec:model_eff}
In order to study the impact of equivariance on model efficiency, we take inspiration from \cite{vecnet} and define a metric `ant factor v2'$=[(1-AUC)\cdot\mathrm{N}_\mathrm{parameters}]^{-1}$ that captures the trade-off between model size and accuracy. Considering this figure of merit, it appears that at least for these particle physics problems, equivariance alone is not the best way to achieve model efficiency. EFP has by far the best ant factor v2, perhaps indicating that constructing a smart basis of input features that contains most or all of the information present in the raw data in a concise and orthogonal manner is the best way to build a highly efficient model. While EFPs have a basis that can be chosen to be equivariant to Lorentz boosts in z-direction (the direction of the beamline), given the performance of other equivariant models it is not possible to attribute the impressive model efficiency of EFPs to equivariance. 

LGN and VecNet.2 are the second and third most efficient models respectively. While LGN is Lorentz equivariant, VecNet.2, which was optimized for model efficiency, is only partially equivariant; this configuration of VecNet includes 4 unconstrained channels. Thus, the relationship between equivariance and model efficiency remains unclear; it appears, as with generalizability, to depend at least somewhat on the specific method of enforcing equivariance.

We also note that as with generalizability, enforcing equivariance is far from the only approach to building efficient GNNs. Network pruning has been successfully applied to GNNs to reduce the number of network parameters required to achieve a certain model accuracy (and thereby improve model efficiency as measured by the ant factor) \cite{gnn_pruning}. Indeed, enforcing equivariance may be seen as a short-cut to a network that has a winning lottery ticket, while an unconstrained network could discover a better, equally sparse lottery ticket, as discussed in \cite{frankle2019lottery}.

\subsection{Data Efficiency}
Data efficiency, in terms of the amount of data needed to train a particular model to a particular accuracy, seems to be the most demonstrable benefit of equivariant models. As shown in Table \ref{tab:data_eff} the Lorentz equivariant top jet tagger LorentzNet achieves higher accuracy and AUC than ParticleNet when trained on very small fractions of the original training dataset; the performance gap increases as the fraction of training data becomes smaller. This trend is seen in other applications of equivariant GNNs to physics data. For example, NequIP, an E(3) equivariant GNN for learning interatomic potentials, is able to achieve SotA accuracy after training on only 1000 graphs \cite{nequip}. The authors are able to attribute this data efficency directly to the equivariance NequIP by comparing predictive error as a function of training set size for equivariant and non-equivariant versions of the same model. 

\begin{table}[]
\begin{tabular}{@{}lllll@{}}
\toprule
Training \% & Model & Accuracy & AUC & $1/\epsilon_B$ \\ \midrule
\multirow{2}{*}{0.5\%} & ParticleNet & 0.913 & 0.9687 & $199\pm 14$ \\ 
 & LorentzNet & \textbf{0.932} & \textbf{0.9793} & \textbf{562} $\pm$ \textbf{72} \\ \hline
\multirow{2}{*}{1\%} & ParticleNet & 0.919 & 0.9734 & $287\pm 19$ \\
 & LorentzNet & \textbf{0.932} & \textbf{0.9812} & \textbf{697} $\pm$ \textbf{58} \\ \hline
\multirow{2}{*}{5\%} & ParticleNet & 0.931 & 0.9807 & $609\pm 35$ \\
 & LorentzNet & \textbf{0.937} & \textbf{0.9839} & \textbf{1108} $\pm$ \textbf{84} \\ \bottomrule
\end{tabular}
\caption{Performance comparison between ParticleNet and LorentzNet by fraction of training data used. Results are averaged on 6 runs. Reproduced from \cite{lorentz_net}\label{tab:data_eff}}
\end{table}

\paragraph{Computational Efficiency} Although data efficiency is highly desirable for minimizing the computational resources required during model training, in many physics use cases computational efficiency during inference is also extremely important. In particle physics in particular, many models need to be used in the online triggering system that must select data for further offline analysis with low microsecond latency \cite{trigger_req}. 

One might naively expect that the model efficiency associated with equivariant models would lead to smaller and therefore more computationally efficient models. Table \ref{tab:topjet_timing} compares the inference times for different top jet tagging models on CPUs and GPUs. The equivariant models LGN and LorentzNet have substantially larger inference times on both devices compared to the non-equivariant models, even when the non-equivariant models are larger (with the exception of LorentzNet and ResNeXt on GPU). Of course factors beyond model size and equivariance contribute to inference time, such as graph construction and the way equivariance is enforced in the model. 

We can investigate the impact of equivariance on computational efficiency more directly by studying the inference time of the tracking models which use the same graph construction method and the same number and size of hidden layers. As seen in Table \ref{tab:tracking_timing}, the non-equivariant InteractionNet has a lower inference time than EuclidNet on CPU while EuclidNet is nearly twice as fast as InteractionNet on GPU. It is thus difficult to establish a clear relationship between equivariance and computational efficiency.

\begin{table*}[]
\centering
\begin{tabular}{@{}lllll@{}}
\toprule
Model       & Equivariance                & Time on CPU (ms/batch) & Time on GPU (ms/batch) & $\mathrm{N}_{\mathrm{parameters}}$ \\ \midrule
ResNeXt     & none                        & 5.5                    & 0.34                   & 1.46M                             \\
PFN         & none                        & \textbf{0.6}           & \textbf{0.12}                  & 82k                               \\
ParticleNet & none                        & 11.0                   & 0.19                   & 498k                              \\
LGN         & $\mathrm{SO}^+{\mathrm{(1,3)}}$ & 51.4                   & 1.66                   & 4.5k                              \\
LorentzNet  & $\mathrm{SO}^+{\mathrm{(1,3)}}$ & 32.9                   & 0.34                   & 224k      \\ \bottomrule                    
\end{tabular}
\caption{Inference time of different SotA and near SotA top taggers. Reproduced from \cite{lorentz_net}. Models are executed on a cluster with an Intel Xeon CPU E5-2698 v4 and an Nvidia Tesla V10032GB. All the inference times are collected with a batch size of 100.\label{tab:topjet_timing}}
\end{table*}

\begin{table*}[]
\centering
\begin{tabular}{@{}lllll@{}}
\toprule
Model          & Equivariance & Time on CPU (ms/batch) & Time on GPU (ms/batch) & $\mathrm{N}_{\mathrm{parameters}}$ \\ \midrule
InteractionNet & none         & \textbf{$\mathbf{4.714 \pm 0.265}$}      & $0.403 \pm 0.043$      & 4392                              \\
EuclidNet      & SO(2)        & $4.866\pm 0.402$       & \textbf{$\mathbf{0.151\pm 0.005}$}       & 2580   \\
\bottomrule
\end{tabular}
\caption{Inference time of two GNN based tracking models. Models are executed on an AMD EPYC 7713 64-Core CPU and an Nvidia A100 40GB GPU. All the inference times are collected with a batch size of 100 over 300 independent runs. \label{tab:tracking_timing}}
\end{table*}

\section{The Path Forward}
\label{forward}
The studies discussed in Section \ref{sec:evaluating} demonstrate that the relationship between equivariance and model performance for physics tasks is more complex than one might initially believe. In this section we discuss future research directions that the physics and ML communities may pursue to better understand the utility of equivariant models and help ensure that the best possible models are developed for important physics tasks.

First, effort should be devoted to isolating what, if any, performance benefits are attributable directly to the enforcement of equivariance in models versus other design choices. Many equivariant models developed for physics tasks employ different methods of enforcing equivariance, different graph construction methods, and different optimization goals, among other cosiderations, making it difficult to compare performance `apples-to-apples'. 

The authors of \cite{lorentz_net} perform an ablation study comparing the performance of LorentzNet and a corresponding non-equivariant version of the model (reproduced in Table \ref{tab:ablation}); LorentzNet achieves a higher accuracy and AUC. This gives some evidence that in this particular instance the model performance improvement may be attributable to equivariance, however, there are other small differences between the two models like the explicit inclusion of the euclidean distance between pairs of particles in the message construction, a feature directly related to the interaction between particles (and one the authors mention as important to LorentzNet performance earlier in the paper). As a first step, we recommend all studies on the application of equivariant models to physics include a carefully constructed ablation study.

\begin{table}[]
\begin{tabular}{@{}llll@{}}
\toprule
Model & Accuracy & AUC & $\epsilon_B^{-1}$ \\
\midrule
LorentzNet (w/o) & 0.934 & 0.9832 & $1105\pm 59$ \\
LorentzNet & \textbf{0.942} & \textbf{0.9868} & $\mathbf{2195\pm 173}$ \\
\bottomrule
\end{tabular}
\caption{Performance comparison between LoretnzNet and a non-equivariant version (LoretnzNet w/o). Reproduced from \cite{lorentz_net}. \label{tab:ablation}}
\end{table}

Without a more systematic study, however, it is impossible to understand in which cases enforcing equivariance will yield more accurate models. The VecNet architecture combined with a hyperparameter importance study is a useful starting point for isolating the impact of equivariance in models. In the case of Lorentz equivariance for top tagging, the VecNet authors found that the most impactful hyperparameters for model AUC were the sizes of the equivariant and non-equivariant channels, and that all of the highest performing models in terms of AUC and model efficiency included both equivariant and non-equivariant channels \cite{vecnet}. Similar studies should be repeated for different tasks, symmetries, and, critically, methods of enforcing equivariance; the results summarized in Section \ref{sec:evaluating} indicate that different approaches to incorporating equivariance may yield substantially different performance. Additionally, the underlying architecture (i.e. CNN vs GNN) may contribute to the expressive ability of the model. 

The results of \cite{vecnet} also suggest that soft- or semi-equivariant models should be further studied for physics tasks. This is a promising direction of research for two reasons. First, as discussed in \cite{approx_symmetry_yu}, although the fundamental theories of physics may obey many symmetries, the real-world data from physics experiments may not be perfectly symmetric due to noise or symmetry-breaking arising from the measuring or data collection apparatus. Thus, models that are biased towards, but not strictly beholden to, a symmetry may be better able to represent these approximate symmetries. Second, it is not established that equivariant models are equivalently expressive as architecturally similar unconstrained models. While the common narrative is that finding an optimal solution in a constrained search space should be an easier task, it is also possible that a constrained model may struggle to escape local minima or that the number of approximately equivalent loss minima may be reduced. 

There are several promising approaches to building soft- or semi-equivariant models that should be studied further. These include mixtures of equivariant and non-equivariant information, as in VecNet \cite{vecnet}, relaxation of equivariance constraints through weighted combinations of equivariant kernels \cite{approx_symmetry_yu} or the addition of non-stationary filters \cite{van2022relaxing}, models constrained to be equivariant only over a subset of a group of transformations \cite{subgroup_equivariance}, and residual pathway priors that allow unconstrained layers to learn from equivariant layers \cite{rpps}.

While equivariance has been a popular method for including physics-based inductive biases into ML models, the analysis in this paper underscores the importance of continued investigation into other approaches to incorporating physics knowledge. For instance, the authors of \cite{lemos2022rediscovering} demonstrate that by a careful construction of the learning task and graph, a GNN can automatically discover the governing equations and properties of real physical systems from limited observations. The authors incorporate Newton's Laws directly into the GNN architecture by setting opposite direction edge weights between the same graph nodes (representing astrophysical objects) to the negatives of each other (Newton's third law) and equate the edge weights to the force on an object, using the equation $F=ma$ (Newton's second law) to calculate the object's acceleration, which is compared to the true acceleration to calculate the network's loss function. Similarly, the authors of \cite{rubanova2021constraint} show that it is possible to build accurate and efficient physical system simulators by tasking a GNN with learning a constraint function that describes which next states are physically disallowed and then solving the optimization problem defined by the learned constraint. Hamiltonian \cite{hamiltonian} and Lagrangian \cite{lagrangian} neural networks have also been shown to improve the modeling of physical characteristics like conservation laws. 

We believe more effort should be invested into these and similar approaches that use physics principles to change the predictive task of a model or the data representation, rather than constraining the optimization space. This should be coupled with systematic studies, as described above, to ascertain the generalizable benefits of these approaches. 

\section{Conclusions}
\label{conclusions}
When building ML models for real-world tasks, it is essential to have a clear picture of the intended goals and evaluation metrics in order to decide what architecture to use. Different domains may have vastly different requirements; in some applications, like real-time inference for autonomous vehicles or the online trigger system at the LHC, model efficiency may be the most important consideration, while in other cases optimal accuracy may be essential. There is no `one size fits all' solution to building the optimal model for a specific task.

In particular, we have shown that even a single model design choice, enforcing equivariance, does not provide universal benefits. While in some cases equivariance may help achieve certain goals like accuracy, generalizability, and data and model efficency, this is not guaranteed. Moreover, equivariance is certainly not the only way to build models that may embody these goals. While equivariant models generally demonstrate generalizability, this can also be achieved through data augmentation, and if the cost of augmentation is low, an unconstrained model may ultimately perform better. Similarly, models can be made more efficient while maintaining accuracy through pruning \cite{lottery_ticket} or parameter quantization \cite{gnn_fpga}.

Additionally, it is worth considering if inductive biases are even necessary to achieve SotA performance on physics tasks. Recently, transformer-based architectures have been utilized with great success in particle physics; in particular, Particle Transformer (ParT) achieve higher accuracy than any other model discussed in this paper on the top-jet tagging task \cite{particle_transformer}. Although it has not yet been studied, it is also possible that transformers, like Interaction Networks, may learn approximately equivariant functions without explicit constraints, and thus generalize effectively as well. However, while transformers may achieve SotA accuracy, they are often very highly parameterized (indeed, versions ParT range from 2M to 100M parameters) and are thus unlikely to be model efficient without additional pruning or other modifications. Again the intended use of the model and selected evaluation metrics are essential considerations when selecting the most appropriate model for a given task. 

Moving forward, more systematic and fewer one-off studies should be conducted in order to understand the impact of equivariance and other inductive biases in real-world applications of ML. We recognize that this likely necessitates a shift in ML publication norms, however we believe it is essential to developing a more complete scientific theory of ML and to maximizing the benefits and utility of applying ML to the physical sciences.

\bibliography{references}
\bibliographystyle{icml2023}

\end{document}